\documentclass[sigconf]{acmart}

\def\BibTeX{{\rm B\kern-.05em{\sc i\kern-.025em b}\kern-.08emT\kern-.1667em\lower.7ex\hbox{E}\kern-.125emX}}

\usepackage{amsfonts, amsmath, amsthm, amssymb, bm, mathtools} 
\usepackage{algorithm,algorithmic}
\usepackage{todonotes}
\usepackage{multirow}

\DeclareMathOperator*{\argmax}{arg\,max}

\copyrightyear{2019} 
\acmYear{2019} 
\setcopyright{acmlicensed}
\acmConference[KDD '19]{The 25th ACM SIGKDD Conference on Knowledge Discovery and Data Mining}{August 4--8, 2019}{Anchorage, AK, USA}
\acmBooktitle{The 25th ACM SIGKDD Conference on Knowledge Discovery and Data Mining (KDD '19), August 4--8, 2019, Anchorage, AK, USA}
\acmPrice{15.00}
\acmDOI{10.1145/3292500.3330932}
\acmISBN{978-1-4503-6201-6/19/08}

\begin{document}
\fancyhead{}
%
\title{Sequential Anomaly Detection using Inverse Reinforcement Learning}

%
\author{Min-hwan Oh}
\affiliation{%
  \institution{Columbia University}
  \city{New York}
  \state{New York}
  \postcode{10027}
}
\email{m.oh@columbia.edu}

\author{Garud Iyengar}
\affiliation{%
  \institution{Columbia University}
  \city{New York}
  \state{New York}
  \postcode{10027}
}
\email{garud@ieor.columbia.edu}


%
\renewcommand{\shortauthors}{  }

%
\begin{abstract}
One of the most interesting application scenarios in anomaly detection is when sequential data are targeted. For example, in a safety-critical environment, it is crucial to have an automatic detection system to screen the streaming data gathered by monitoring sensors, and to report abnormal observations if detected in real-time. Oftentimes, stakes are much higher when these potential anomalies are intentional or goal-oriented.
We propose an end-to-end framework for sequential anomaly detection using inverse reinforcement learning (IRL), whose objective is to determine the decision-making agent's underlying function which triggers his/her behavior. The proposed method takes the sequence of actions of a target agent (and possibly other meta information) as input. The agent's normal behavior is then understood by the reward function which is inferred via IRL. 
We use a neural network to represent a reward function. Using a learned reward function, we evaluate whether a new observation from the target agent follows a normal pattern. In order to construct a reliable anomaly detection method and take into consideration the confidence of the predicted anomaly score, we adopt a Bayesian approach for IRL. The empirical study on publicly available real-world data shows that our proposed method is effective in identifying anomalies.
\end{abstract}

%
%

%

%

%
\maketitle

\section{INTRODUCTION}
Anomaly detection, or outlier detection refers to (automatic) identification of unforeseen or abnormal phenomena embedded in a large amount of normal data. 
However, anomaly detection is a challenging topic, mainly because of the insufficient knowledge and inaccurate representative of the so-called anomaly for a given system.  Often, there are no examples from which distinct features of the anomaly could be learned. 
Furthermore, in many cases, even when the available data is large, there is the limited availability of labels to model anomaly detection as a discriminative classification task. In most common practices, one must rather learn a precise generative model of normal patterns and detect anomalies as cases that are not sufficiently explained by this model.

One of the most interesting application scenarios in anomaly detection is when sequential data are targeted. For example, in a safety-critical environment, it is crucial to have an automatic detection system to screen the streaming data gathered by monitoring sensors, and to report abnormal observations if detected in real-time. Oftentimes, stakes are much higher when these potential anomalies are intentional or goal-oriented.
Various types and large amounts of sequential data are available due to increasing pervasiveness of mobile devices, surveillance systems or other monitoring devices, which pose new challenges both computationally and statistically, and thus require novel approaches in discovering useful patterns.
As opposed to static anomalies which are objects that exhibit abnormal behavior at a single snap shot of time, sequential anomalies are objects or instances that exhibit abnormal behavior over several time periods (and can potentially be intentional). Hence, detecting sequential anomalies is more complex, as one needs to capture the objects or instances that deviate from normal behavior by establishing continuity of actions in potentially multiple dimensions


We propose an end-to-end framework for sequential anomaly detection using inverse reinforcement learning (IRL). The objective of IRL is to determine the decision making agent's underlying reward function from its behavior data \citep{russell1998learning}. A reward function incentivizes an agent to act in a certain way, and hence describes the preferences of the agent whose objective is to collect as much reward as possible (see Section \ref{sec:prelim} for more detail). The significance of IRL has emerged from problems in diverse research areas. In robotics \citep{argall2009survey}, IRL provides a framework for making robots learn to imitate the demonstrator's behavior using the inferred reward function. In human and animal behavior studies \citep{montague2002neural}, the agent's behavior could be understood by the reward function since the reward function reflects the agent's objectives and preferences. 

The proposed method takes the trajectories of target agents as input. In addition to the sequence of coordinates, our method can also incorporate other meta information, e.g. date, time,  and possibly the features of landmarks if available, as additional input. The agent's normal behavior is then understood by the reward function which is inferred via inverse reinforcement learning. 
We use a neural network to represent a reward function for each target agent. Using a learned reward function, we can evaluate whether a new observation from the target agent follows a normal pattern. In other words, if the new observation gives a low reward, then it implies that the observation is not explained by the preferences of the agent that we have learned so far, and that it can be considered as a potential anomaly.

In order to construct a reliable anomaly detection method, we need to also take into consideration the confidence of the predicted anomaly score.
Rather than blindly taking the estimated reward as a normality score only, we also consider model confidence of the predicted values. This is crucial especially when false alarms can incur costs (which is typical in many applications). For this reason, we consider a Bayesian approach for IRL in this work. Bayesian IRL formulates the reward preference as the prior and the behavior compatibility as the likelihood, and find the posterior distribution of the reward function \citep{ramachandran2007bayesian}. Most of the existing IRL methods with non-linear function approximation are in non-Bayesian (point estimation) settings \citep{wulfmeier2015maximum,finn2016guided,finn2016connection}; hence cannot incorporate model uncertainty. 
There are a number of advantages of Bayesian approaches in learning perspectives: We do not need a completely specified optimal policy as input to the IRL agent, nor do we need to assume that the trajectories from the target agent has no anomalies, hence allowing a small portion of anomalies to be present in the trajectories when learning the normal behavior; Also, we can incorporate external information about specific IRL problems into the prior of the model. 
Furthermore, with model confidence at hand we can treat uncertain inputs and special cases explicitly. For example, the reward function might return a value with high uncertainty. In this case we might decide to pass the input to a human for validation.

The main contributions of this paper are as follows:
\begin{itemize}
    \item To the best of our knowledge, this is the first time that IRL is used for anomaly detection, which is a natural way of understanding the underlying motive of observed behaviors when anomalous actions are potentially intentional or goal-oriented.
    \item We incorporate the model uncertainty in the IRL problem in a very simple and scalable manner, which may be of independent interest. This model uncertainty is shown to help identifying an anomaly.
    \item Our method can incorporate trajectories with varying lengths as input and can be used for real-time detection.
    \item The empirical study on publicly available real-world data shows that our proposed method is effective in identifying anomalies.

\end{itemize}
The organization of the paper is as follows:
In Section 2, we discuss literature in sequential anomaly detection. Note that we defer the discussion (and literature review) on IRL entirely to Section 3 due to the limited space. Section 3 gives a brief overview of IRL and its Bayesian extension. Section 4 presents our algorithm and anomaly detection procedure. Section 5 discusses the empirical study of our proposed method. Finally, Section 6 concludes our paper.


\section{Related Work}

Anomaly detection has its roots in the more general problem of outlier detection. In many cases the data is static, rather than evolving over time. There are a number of different methods available for outlier detection, including supervised approaches \citep{abe2006outlier}, distance-based \citep{angiulli2009dolphin,knorr2000distance}, density-based \citep{breunig2000lof}, model-based \citep{he2003discovering}
and isolation-based methods \citep{liu2008isolation}.

Sequential anomaly detection methods constitute an important category of anomaly detection methods, which usually uses object-trajectories of tracked persons, or cars in traffic scenes for example 
\cite{piciarelli2008trajectory, sillito2008semi, hu2006system, li2013visual,  dee2004detecting}, but also temporally shorter and dense trajectory representations, such as representative trajectories for crowd flow obtained from clustering particle trajectories in \cite{wu2010chaotic}.  \cite{sillito2008semi} presents a method that uses cubic spline curves to parametrize trajectories and an incremental one-class learning approach using Gaussian mixture models.  \cite{li2013visual} introduces trajectory sparse reconstruction analysis (SRA) that constructs a normal dictionary set which is used to reconstruct test trajectories.
In \citep{lee2008trajectory}, a trajectory is split into various partitions (at equal intervals) and a hybrid of distance and density based approaches is used to classify each partition as anomalous or not. In \citep{ge2010top}, the authors compute a score based on the evolving moving direction and density of trajectories, and make use of a decay function to include previous scores. \cite{bu2009efficient} present a method for monitoring anomalies over continuous trajectory streams by using the local continuity
characteristics of trajectories to cluster trajectories on a local level; anomalies are then identified by a pruning mechanism. 
However, many of these previous methods require feature engineering to perform analysis or preprocessing of the data.


There is also a line of work that utilize clustering approaches.
\cite{hu2006system} uses a hierarchical clustering of trajectories depending on spatial and temporal information and a chain of Gaussian distributions to represent motion patterns.
In \cite{piciarelli2008trajectory}, single-class support vector machine (SVM) clustering is used to identify anomalous trajectories in traffic scenes.
\cite{wang2018detecting} distinguish regular trajectories and anomalous trajectories by applying adaptive hierarchical clustering based on an optimal number of clusters.
However, many of these methods are designed to be used as batch-based method, and cannot be used in real-time in online settings.


Recent advances in artificial neural network and availability of larger datasets allowed the use of deep learning in the domain of sequential anomaly detection. \cite{sabokrou2018deep} uses convolutional neural network (CNN) transferred from a pre-trained supervised network and ensures the detection of (global) anomalies in video scenes. However, this is only limited to a fixed location (fixed camera view point).
Recurrent neural network (RNN) based approaches have been proposed for anomaly detection \cite{malhotra2016lstm,oshea2016recurrent}. These approaches learn a model to reconstruct the normal data (e.g. when a system is in perfect health) such that the learned model could reconstruct the sub-sequences which belong to normal behavior. The learned model leads to high reconstruction error for anomalous sub-sequences, since it has not seen such data during training. However, their evaluation uses either only univariate time series or time series with more regularity such as bounded sequences and periodic patterns. Also, these method typically requires the fixed length of sequences (or fixed window-size). 

 The majority of methods discussed so far do not necessarily address the scenario where the target agent may be intentionally malicious or goal-directed. \cite{dee2004detecting} addresses this issue and uses inexplicability scores to measure the extent to which a trajectory can be regarded as goal-directed. While our work and \cite{dee2004detecting} share a similar motivation, their method however still requires feature-engineering and limited to be used in a online setting.

There are also other work which use (forward) reinforcement learning to detect anomalies \cite{lu2017motor} or certain patterns of interest  \cite{kanezashi2018adaptive} in dynamic systems. However, these methods require the predefined notion of reward signals whereas our approach is to infer such reward functions since it is unknown.



\section{Preliminaries}\label{sec:prelim}

\subsection{\textit{Forward} and \textit{Inverse} Reinforcement Learning (RL) Basics}
We assume that the environment is modeled as a Markov decision process (MDP) $\left< S, A, T, R, \gamma, p_0 \right>$ where $S$ is the finite set of states; $A$ is the finite set of actions; $T(s, a, s')$ is the state transition probability of changing to state $s'$ from state $s$ when action $a$ is taken; $r(s, a)$ is the immediate reward of executing action $a$ in state $s$; $\gamma \in [0, 1)$ is the discount factor; $p_0(s)$ denotes the probability of starting in state $s$. For notational convenience, we use the vector $r = \left[r_1, ..., r_D\right]$ to denote the reward function.

A policy is a mapping $\pi : S \rightarrow A$. The value of policy $\pi$ is the expected discounted return of executing the policy, defined as $V^\pi = \mathbb{E}[\sum_{t=0}^\infty \gamma^t r(s_t, a_t)|p_0, \pi]$. 
The value function of policy $\pi$ for state $s$ is computed by $V^\pi(s) = r(s,\pi(s)) + \gamma \sum_{s'\in S} T(s,\pi(s),s')V^\pi (s')$ so that the value is calculated by $V^\pi = \sum_{s\in S} p_0(s) V^\pi(s)$.
Given an MDP, the agent's objective is to execute an optimal policy $\pi^*$ that maximizes the value function for all the states, which should satisfy the Bellman optimality equation: $V^*(s) = \max_{a\in A}[r(s,a) + \gamma \sum_{s'\in S} T(s,a,s')V^* (s')]$.

In (forward) RL, $\pi^*$ is the policy to be learned. In IRL on the other hand, the reward function is not explicitly given. Hence, the goal of IRL is to learn a reward function $r^*(\cdot)$ that that explains the demonstrator's behavior:
\begin{equation*}
\mathbb{E}\left[\sum_{t=0}^\infty \gamma^t r^* (s_t, a_t) | \pi^*\right] \geq \mathbb{E}\left[\sum_{t=0}^\infty \gamma^t r^* (s_t, a_t) | \pi \right],   \quad \forall \pi
\end{equation*}
where $\pi^*$ is the demonstrator's policy which is optimal (or near-optimal) with respect to the unknown reward function $r^*(s_t, a_t)$. 

However, the optimal policy $\pi^*$ is not always explicitly given in most cases. Instead, demonstrations or samples from $\pi^*$ are given. In other words, we assume that the agent's demonstration is generated by executing an optimal policy $\pi^*$ with some unknown reward function $R^*$, given as the set $\mathcal{T}$ of $M$ trajectories where the $m$-th trajectory is a sequence of state-action pairs: $\tau_m = \{ (s_{m,1}, a_{m,1}), (s_{m,2}, a_{m,2})..., (s_{m,H}, a_{m,H})) \}$ where $H$ is a horizon length, which can vary from a trajectory to another. 
Also, we sometimes overload the term $\pi$ and denote the trajectory-wise policy as $\pi(\tau)$ to represent a policy being followed to generate an entire trajectory following $\pi(s)$ sequentially.

\subsection{Maximum Entropy IRL}\label{sec:maxEnt}
Maximum entropy IRL framework \citep{ziebart2008maximum} models the demonstrations using a Boltzmann distribution, where the energy is given by the trajectory-wise reward function $R (\tau | \theta)$:
\begin{equation} \label{maxEntEq}
p (\tau | \theta) = \frac{1}{Z} \exp(R (\tau | \theta))
\end{equation}
where $R (\tau | \theta) = \sum_{(s,a) \in \tau} r_\theta (s, a)$ and $r_\theta (s, a)$ is a learned reward function parametrized by $\theta$, and the partition function $Z$ is the integral of $\exp(R (\tau | \theta))$ over all trajectories consistent with the MDP's dynamics. Under this model, trajectories with equal reward are assigned the same likelihood, and higher-reward trajectories are exponentially more preferred by the demonstrator. The parameters $\theta$ are chosen to maximize the likelihood of the (given) demonstrated trajectories: 
\begin{align*}
\mathcal{L}(\theta) &= \frac{1}{M} \sum_{m=1}^M \log p (\tau_m | \theta)\\
&= \frac{1}{M} \sum_{m=1}^M R (\tau_m | \theta) - \log \sum_\tau \exp(R(\tau | \theta)) 
\end{align*}
Computing the gradient with respect to $\theta$ gives the following:
\begin{align}
\nabla_\theta \mathcal{L} = \mathbb{E}_{\tau_m \sim \pi^*} \left[ \nabla_\theta R(\tau_m | \theta) \right] - \mathbb{E}_{\tau \sim \pi_{\theta} } \left[ \nabla_\theta R(\tau | \theta) \right].\label{gradLoss}
\end{align}
Here $\pi^*$ is the demonstrator's policy and $\pi_{\theta}$ is a (soft) optimal policy under reward parameter $\theta$ \citep{ziebart2008maximum}. This derivation is applicable to any differentiable reward functions. \cite{ziebart2008maximum} use linear reward functions in their work, and they show that the gradient implies that the optimal policy under $\theta^*$ matches the feature expectation of the demonstrator's policy. 

Computing the second expectation in (\ref{gradLoss}) requires finding the soft optimal policy under the current reward parameter $\theta$ and computing its expected state visitation frequencies using a variant of the value iteration algorithm for (forward) RL. However, for large or continuous domains, this becomes intractable, since this computation scales exponentially with the dimensionality of the state space. This method also requires repeatedly solving an MDP in the inner loop of an iterative optimization, further increasing the computational difficulty of larger systems. \cite{finn2016guided, finn2016connection} extend maximum entropy IRL to sample-based IRL where dynamics of MDP is not given and a reward function is parametrized with non-linear function, e.g. neural networks; hence in order to estimate $Z$, they generate background sample trajectories (see \cite{finn2016guided} for more detail). The sample-based IRL \cite{finn2016guided} is closely related to other sample-based maximum entropy methods, including relative entropy IRL by \cite{boularias2011relative} and path integral IRL by \cite{kalakrishnan2013learning}, which can also handle unknown dynamics. However, unlike these prior methods, \cite{finn2016guided} adapt the sampling distribution using policy optimization.

\subsection{Bayesian framework for IRL}\label{sec:bayesianIRL}
The main idea of Bayesian IRL (BIRL) is to use a prior to encode the reward preference and to formulate the compatibility with the demonstrator's policy as a likelihood in order to derive a probability distribution over the space of reward functions, from which the demonstrator's reward function is extracted \citep{ramachandran2007bayesian}.
Assuming that the reward function entries are independently distributed, the prior is defined as
\begin{equation*}
    P(r) = \prod_{(s,a) \in \mathcal{T}} P\left( r(s,a) \right).
\end{equation*}
Various distributions can be used as the prior. For example, the uniform prior can be used if we have no knowledge about the reward function other than its range, and a Gaussian or a Laplacian prior can be used if we prefer rewards to be close to some specific values.
The likelihood in BIRL is defined as an independent exponential distribution analogous to the softmax function:
\begin{align*}
    P(\mathcal{T} | r) &= \prod_{\tau \in \mathcal{T}} \prod_{(s,a)\in \tau} P(a | s, r)
\end{align*}
The posterior over the reward function is then formulated by combining the prior and the likelihood,
using Bayes theorem:
\begin{equation*}\label{eq:birl}
    P(r|\mathcal{T}) \propto P(\mathcal{T}|r)P(r) 
\end{equation*}
The reward function can be inferred from the model by computing the posterior mean using a Markov chain Monte Carlo (MCMC) algorithm \citep{ramachandran2007bayesian}.

\subsubsection*{MAP Inference}
\citep{choi2011map} show that using the posterior mean may not be a good idea since it may yield a reward function whose corresponding optimal policy is inconsistent with the demonstrator's behaviour. In other words, the posterior mean integrates the error over the entire space of reward functions by including (possibly) infinitely many rewards that induce policies that are inconsistent with the demonstration trajectories. Instead, the maximum-a-posteriori (MAP) estimate could be a better solution for IRL. Then IRL can be formulated as a posterior optimization problem and \citep{choi2011map} propose a gradient method to calculate the MAP estimate that is based on the (sub)differentiability of the posterior distribution. Most of the non-Bayesian IRL algorithms in the literature \citep{ng2000algorithms, ratliff2006maximum, neu2012apprenticeship, ziebart2008maximum, syed2008apprenticeship} can be cast as searching for the MAP reward function in BIRL with different priors and different ways of encoding the compatibility with the demonstrator's policy.

Note that both BIRL \cite{ramachandran2007bayesian} and MAP framework for IRL \cite{choi2011map} have been limited to tabular settings, i.e. without function approximation. Hence, it cannot be directly applied to our setting. In the next section, we introduce our method and discuss how we can combine the notion of uncertainty with sampled-based IRL.

\section{METHOD }
Our proposed method extends the sample-based maximum entropy IRL \cite{finn2016guided} to Bayesian framework to approximate a distribution over reward functions. The main idea of the proposed method is the following. We first assume the parameters of the reward function come from a prior distribution. At the start of each training iteration, we sample a single reward function from its approximate posterior. We then follow a sample generating policy for the duration of the iteration to generate background trajectories (recall that we need background sample trajectories to estimate the partition function $Z$ in Eq. (\ref{maxEntEq}) since we do not know the underlying dynamics of MDP), and improve the policy with respect to the sampled reward function. 
Then, using the generated background trajectories along with the demonstrated trajectories from the target agent, we estimate the gradient and update the (sampled) reward function. We repeat this process until convergence.

We parametrize our reward functions as neural networks, expanding their expressive power without hand-engineered features. Previous work has shown that an affine reward function is not expressive enough to learn complex behaviors \citep{wulfmeier2015maximum, finn2016guided}. 
While the expressive power of nonlinear reward functions provide a range of benefits, they introduce significant model complexity to an already underspecified IRL objective and additionally learning the distribution over this nonlinear reward function adds more complexity.
In the following sections, we present how we approximate the posterior distribution over reward functions and how we can remedy increased computational complexity. Then, we show how we approximate the gradient for the reward function updates. Finally, we present how we operationalize the entire algorithm.

\subsection{Approximate Posterior Distribution}

Many previous literature on Bayesian neural network studied uncertainty quantification founded on parametric Bayesian inference \citep{blundell2015weight,gal2016dropout}. (For more detailed review on Bayesian neural network, see Section \ref{sec:BNN} in the appendix.) In this work, we consider a non-parametric bootstrap of functions. 


\subsubsection{Bootstrap ensemble}\label{bootstrap}

Bootstrap is a simple technique for producing a distribution over functions with theoretical guarantees \cite{bickel1981some, efron1994introduction}. It is also general in terms of the class of models that we can accommodate. In its most common form, a bootstrap method takes as input a data set $\mathcal{D}$ and a function $f_\theta$. We can transform the original dataset $\mathcal{D}$ into $K$ different data sets $\{ \mathcal{D}_k \}_{k=1}^K$'s of cardinality equal to that of the original data $\mathcal{D}$ that is sampled uniformly with replacement.
Then we train $K$ different models. For each model $f_{\theta_k}$, we train the model on the data set $\mathcal{D}_k$. So each of these models is trained on data from the same distribution but on a different data set. Then if we want to approximate sampling from the distribution of functions, we sample uniformly an integer $k$ from 1 to $K$ and use the corresponding function $f_{\theta_k}$. 

\begin{figure}[ht]
    \centering
    \includegraphics[width=0.87\linewidth]{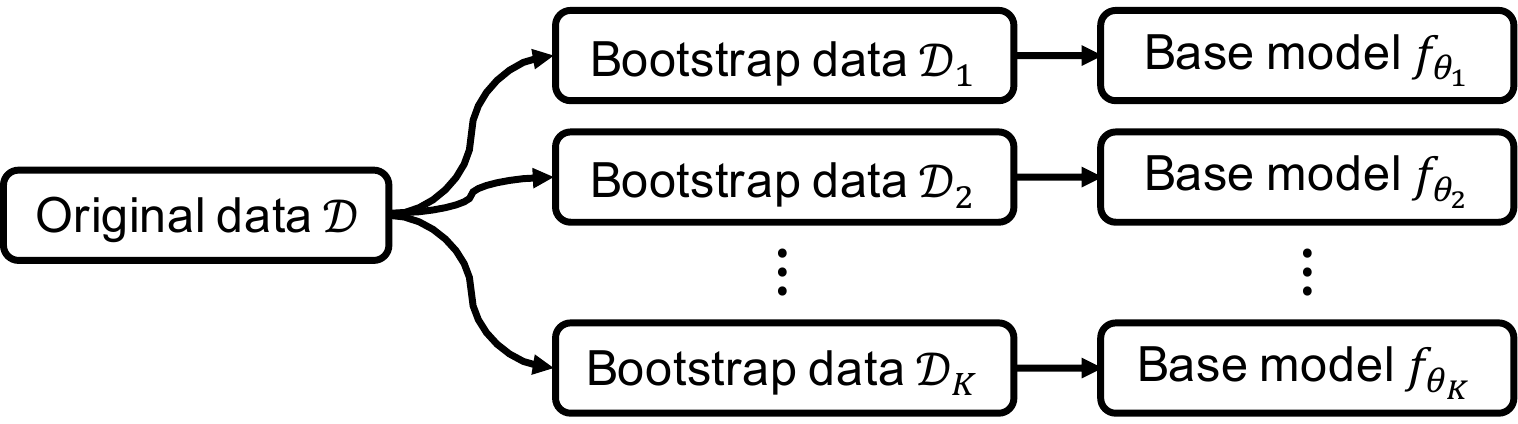}
    \caption{Bootstrap ensemble sampling. each base model is trained on randomly perturbed data }
    \label{fig:bootstrap_ensemble}
\end{figure}

In cases of using neural networks as base models $f_{\theta_k}$, bootstrap ensemble maintains a set of $K$ neural networks $\{ f_{\theta_k} \}_{k=1}^K$ independently on $K$ different bootstrapped subsets of the data. It treats each network as independent samples from the weight distribution. In contrast to traditional Bayesian approaches discussed earlier, bootstrapping is a frequentist method, but with the use of the prior distribution, it could approximate the posterior in a simple manner. Also it scales nicely to high-dimensional spaces, since it only requires point estimates of the weights. However, one major drawback is that computational load increase linearly with respect to the number of base models. In the following section, we discuss how to mitigate this issue and still maintain a reasonable uncertainty estimates.

\subsubsection{Single network with $K$ output heads}

\begin{figure}[]
    \centering
    \includegraphics[width=0.87\linewidth]{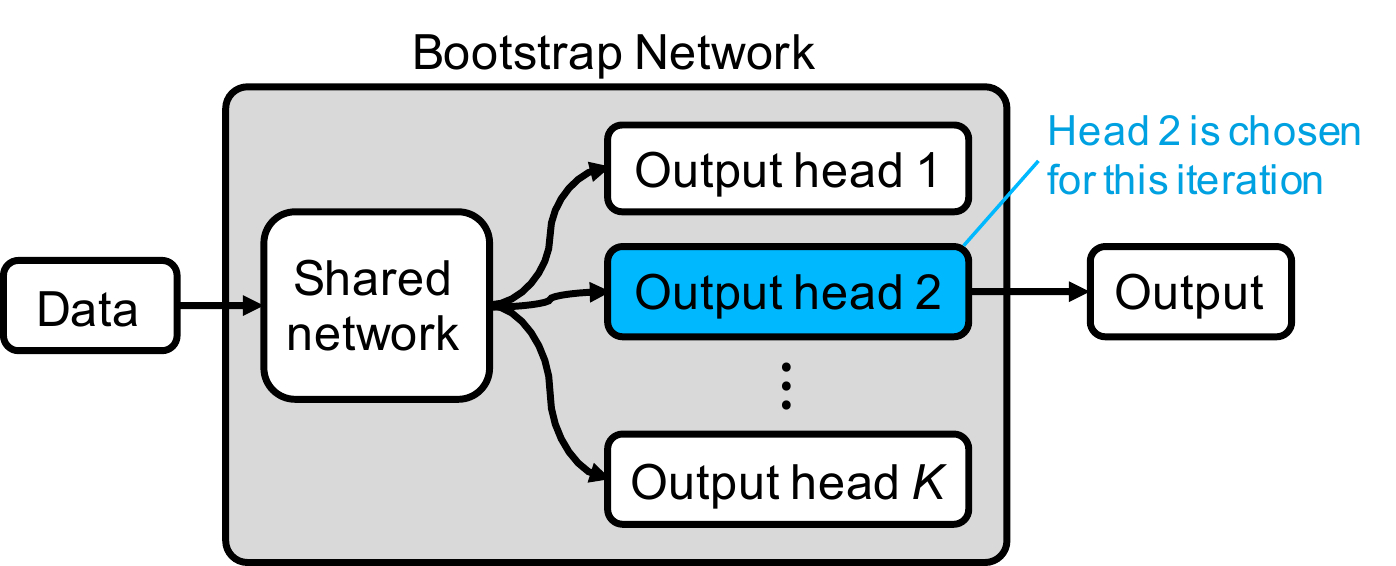}
    \caption{Illustration of bootstrap network with $K$ output heads. In each iteration, an output head is randomly chosen}
    \label{fig:bootstrap_network}
\end{figure}

Training and maintaining a multiple independent neural networks is computationally expensive especially when each base network is a large and deep neural network. In order to remedy this issue, we adopt a single network framework which is scalable for generating bootstrap samples from a large and deep neural network \cite{osband2016deep, oh2019crowd}.
The network consists of a shared architecture with $K$ bootstrapped heads branching off independently (as shown in Figure \ref{fig:bootstrap_network}). Each head is trained only on its bootstrapped sub-sample of the data as described in Section \ref{bootstrap}. The shared network learns a joint feature representation across all the data, which can provide significant computational advantages at the cost of lower diversity between heads. This type of bootstrap can be trained efficiently in a single forward/backward pass.

To capture model uncertainty, we assume a prior distribution over its weights, e.g. with a Gaussian prior $[\theta_s, \theta_1, ..., \theta_K] \sim \mathcal{N}(0, \sigma^2)$ for fixed variance $\sigma^2$, where $\theta_s$ is the parameter of the shared network and $\theta_1, ..., \theta_K$ are the parameters of bootstrap heads $1,.., K$. 
For brevity of notations, we overload the term $\theta_k = [\theta_k, \theta_s]$ since $\theta_s$ is shared across all samples. For each iteration of training procedure, we sample the model parameter $\Hat{\theta}_k \sim q(\theta)$ where $q(\theta)$ is a bootstrap distribution. In other words, at each iteration we randomly choose which head to use to predict a reward score and update the parameters. In the following section, we present how we update the reward parameter $\theta$.

\subsection{Parameter Update}


Suppose we sampled the current reward parameter, i.e. sampled an output head from $K$ heads, and denote the sampled parameter as $\theta$ (instead of $\theta_k$) for the notational brevity in this section. Hence, our reward function is expressed as $r_{\theta}(s,a)$. Following MAP inference arguments in Section \ref{sec:bayesianIRL} and Eq. \ref{eq:birl},
we can then reformulate the IRL problem into the posterior optimization problem, which is finding $\theta_{\text{MAP}}$ that maximizes the (log unnormalized) posterior:
\begin{align*}
    \theta_{\text{MAP}} =  \argmax_{\theta} \mathbb{E}_{\tau \sim \mathcal{D}} [\log  p(\tau | \theta) + \log p(\theta)]
    := \argmax_{\theta} J(\theta) 
\end{align*}
where the distribution $p(\tau | \theta)$ is defined as in Eq. (\ref{maxEntEq}) and $p(\theta)$ is a prior distribution of $\theta$.
We can compute the gradient with respect to $\theta$ as follows:
\begin{align*}
    \frac{\partial}{\partial \theta} J(\theta) &= \mathbb{E}_{\tau \sim \mathcal{D}} \left[  \frac{\partial}{\partial \theta} \log p (\tau | \theta) + \frac{\partial}{\partial \theta} \log p (\theta) \right]\\
    &= \mathbb{E}_{\tau \sim \mathcal{D}} \left[  \sum_{(s,a)\in \tau} \frac{\partial}{\partial \theta} r_{\theta} (s, a) \right] - \frac{\partial}{\partial \theta} \log Z_\theta + \frac{\partial}{\partial \theta} \log p (\theta)
\end{align*}
where $\mathcal{D}$ is the distribution of the observed trajectories. We can then write $\frac{\partial}{\partial \theta} \log Z_\theta$ as $\mathbb{E}_{\tau' \sim \pi}  \left[  \sum_{(s,a)\in \tau'} \frac{\partial}{\partial \theta} r_{\theta} (s, a) \right]$ where $\pi = \pi(\tau)$ is a sample-generating policy discussed in Section \ref{sec:maxEnt}. Now, we approximate the gradient with finite samples:

\begin{align*}
    \frac{\partial}{\partial \theta} J(\theta) &\approx \frac{1}{|\mathcal{T}_d|} \sum_{\tau \in \mathcal{T}_d}    \sum_{(s,a)\in \tau} \frac{\partial}{\partial \theta} r_{\theta} (s, a)\\ 
    &\quad - \frac{1}{|\mathcal{T}_s|}\sum_{\tau_j \in \mathcal{T}_s}  \sum_{(s,a)\in \tau_j} \frac{\partial}{\partial \theta} r_{\theta} (s, a)  + \frac{\partial}{\partial \theta} \log p (\theta)
\end{align*}
where $\mathcal{T}_d$ is a set of demonstrated trajectories from the target agent, and $\mathcal{T}_s$ is a set of background trajectories the policy $\pi$ generates. Now, note that this policy $\pi$ is being improved as we update reward parameters.
Hence, trajectories in $\mathcal{T}_s$ are the samples collected for policy improvement and are also used to update the reward function. However, trajectories sampled from these intermediate policies are biased samples since those are not sampled from the soft optimal policy under the current reward as shown in Eq. (\ref{gradLoss}). To address this issue, we adopt a technique used in \cite{finn2016guided} to use importance sampling to re-weight the samples to make the ones with higher reward more likely (or ones that are unlikely from the current policy more likely).

Define $w_j := \frac{\exp(R (\tau_j | \theta))}{q(\tau_j)}$ where $R (\tau_j | \theta) = \sum_{(s,a) \in \tau_j} r_\theta (s, a)$, and $W := \sum_j w_j$. The gradient is then given by:
\begin{align}
    \frac{\partial}{\partial \theta} J(\theta) &\approx \frac{1}{|\mathcal{T}_d|} \sum_{\tau \in \mathcal{T}_d}    \sum_{(s,a)\in \tau} \frac{\partial}{\partial \theta} r_{\theta} (s, a) \notag\\ 
    &\quad - \frac{1}{W}\sum_{\tau_j \in \mathcal{T}_s}  \sum_{(s,a)\in \tau_j} w_j \frac{\partial}{\partial \theta} r_{\theta} (s, a)  + \frac{\partial}{\partial \theta} \log p (\theta) \label{eq:gradApprox}
\end{align}
Since in our case the reward function $r_\theta(s,a)$ is represented by a neural network, this gradient can be computed efficiently by backpropagating $-\frac{w_j}{W}$ for each background trajectory $\tau_j \in \mathcal{T}_s$ and $\frac{1}{|\mathcal{T}_d|}$ for each demonstration trajectory $\tau \in \mathcal{T}_d$. 
The algorithm alternates between optimizing the reward function $r_\theta(\cdot)$ using this gradient estimate, and optimizing the policy $\pi(\tau)$ with respect to the current reward function. We also use a neural network to represent policy $\pi(\tau)$. For policy optimization, we use trust-region policy optimization (TRPO) \cite{schulman2015trust}, a state-of-the-art policy gradient algorithm.

\subsection{Algorithm Overview}

Suppose we are interested in individual target agent's behavior, hence we learn a single reward function for a given target agent and $\mathcal{T}_d$ contains trajectories from a single target agent. It is important to note that we can also consider learning a reward function for a group of individuals to capture an underlying common behaviors of the group in other applications. Our method can also be used even in that scenario. However, we focus on learning a reward function for a single target agent in this section for the ease of the presentation.

We initialize the reward network parameters $\theta$ and policy network $\pi$, e.g. with a Gaussian distribution with mean $\mu = 0$ and variance $\sigma^2 = 0.1$. We are given a set of trajectories from a target agent $\mathcal{T}_d = \{\tau_1, ..., \tau_M\}$. For each iteration, we sample uniformly at random $k \in \{1, ..., K\}$ to choose an output head $k$, hence a reward function $r_{\theta_k}(\cdot)$. Given this reward function, we follow $\pi(\tau)$ to generate a new set of trajectories $\mathcal{T}_\text{traj}$ which are added to the background samples $\mathcal{T}_s$ to estimate the partition function $Z$ in maximum entropy formulation in Eq. (\ref{maxEntEq}). Using newly updated background samples $\mathcal{T}_s$, we update the reward parameter $\theta_k$ using Eq. (\ref{eq:gradApprox}). Note that the reward function is updated using all samples collected thus far.
Then based on the current reward function, we update the policy $\pi$ using TRPO (or any policy gradient method). Then we repeat this process until convergence. This procedure returns  a learned reward function $r_\theta(\cdot)$ and also a trajectory generating policy $\pi(\tau)$ as a byproduct.
Our algorithm is summarized in Algorithm \ref{algo-sampleBayesianIRL} which uses Algorithm \ref{algo-rewardUpdate} as a sub-routine.


\begin{algorithm}
\caption{IRL with bootstrapped reward}
\begin{algorithmic}[1]
    \STATE Obtain trajectories $\mathcal{T}_d$ from the target agent
    \STATE Initialize policy $\pi$ and reward functions $\{ r_{\theta_k} \}_{k=1}^K$
    \FOR{iterations $n = 1$ to $N$}
        \STATE Sample $r_{\theta_k}$ uniformly at random
        \STATE Generate samples $\mathcal{T}_\text{traj}$ from $\pi$
        \STATE Append samples: $\mathcal{T}_s \leftarrow \mathcal{T}_s \cup \mathcal{T}_\text{traj}$
        \STATE Update $r_{\theta_k}$ using Alogrithm 2 with $\mathcal{T}_s$ and $\theta_k$
        \STATE Update $\pi$ with respect to $r_{\theta_k}$
        using TRPO \cite{schulman2015trust}
    \ENDFOR
    \RETURN optimized reward parameters $\{ \theta_k \}_{k=1}^K$ and policy $\pi$
\end{algorithmic}
\label{algo-sampleBayesianIRL}
\end{algorithm}

\begin{algorithm}
\caption{Reward function update}
\begin{algorithmic}[1]
    \REQUIRE $\theta_k, \mathcal{T}_d, \mathcal{T}_s$ 
    \FOR{iterations $j = 1$ to $J$}
        \STATE Sample background batch $\mathcal{\hat{T}}_s \subset \mathcal{T}_s$
        \STATE Sample demonstration batch $\mathcal{\hat{T}}_d \subset \mathcal{T}_d$
        \STATE Append demonstration to background: $\mathcal{\hat{T}}_s \leftarrow \mathcal{\hat{T}}_s \cup \mathcal{\hat{T}}_d$
        \STATE Estimate $\frac{d\mathcal{L}(\theta_k)}{d\theta_k}$ using $\mathcal{\hat{T}}_d$ and $\mathcal{\hat{T}}_s$
        \STATE Update parameters $\theta$ using gradient $\frac{d\mathcal{L}(\theta_k)}{d\theta_k}$ in Eq. (\ref{eq:gradApprox})
    \ENDFOR
    \RETURN optimized reward parameters $\theta_k$
\end{algorithmic}
\label{algo-rewardUpdate}
\end{algorithm}

\subsection{Normality Score}
Once the reward function is learned on the demonstrated trajectories, we can compute the normality score of a new observation $(s,a)$, which we define as
\begin{equation}\label{eq:normalityScore}
    n(s,a) = \frac{ \Bar{r}_\theta(s,a) - \Bar{r}}{\Bar{\sigma}_r}
\end{equation}
where $\Bar{r}$ and $\Bar{\sigma}_r$ are an average reward and standard deviation for the entire observations, and $\Bar{r}_\theta(s,a) = \frac{1}{K}\sum_k \Bar{r}_{\theta_k}(s,a)$ is the mean reward for the new observation. Note that during training we sample a single output head to compute a gradient and update reward parameters, but in test time prediction, we output from all output heads to compute the mean reward.
Also, it is important to note that a target agent that has trajectories with more regular patterns will have a smaller $\sigma_r$ than ones with more diverse trajectories. This normalization helps balance out this effect when deciding an anomaly. Also we define $N(\tau) := \frac{1}{|\tau|}\sum_{(s,a)\in \tau} n(s,a)$ to be a trajectory-wise normality score. Then, an observation (either state-action pair observation $(s,a)$ or a trajectory $\tau$) can be considered as an anomaly if it has a low normality score, e.g. a normality score is below some threshold $\epsilon$. Now, the exact value of threshold $\epsilon$ depends on specific applications. In some applications, one can allow the threshold to change dynamically depending on time, or other meta information Or, one may use different thresholds for different target agents. Either way, this score indicates how much an observation deviates from the normal score.

\subsection{Incorporating Model Uncertainty when Deciding an Anomaly}\label{sec:incorporateModelUncertainty}
In addition to the predicted reward for a new observation (and the normality score based on it), we also have the predicted variance over the $K$ reward estimates:
\begin{equation*}
    \sigma^2_r(s,a) = \frac{1}{K}\sum_{k=1}^K \left(r_{\theta_k}(s,a) - \Bar{r}_\theta(s,a)\right)^2
\end{equation*}
which represent how uncertain the model is regarding this normality measure. Hence, we can incorporate this information when deciding an anomaly. For example, once normality score finds a potential anomaly with $n(s,a) \leq \epsilon$, one can then put another threshold $\sigma_r(s,a) \leq \gamma$ for some fixed $\gamma$, i.e. the model has to be certain that an observation is an anomaly when assigning an anomaly. This may be useful when false positives (false alarms) incur a higher cost, hence one can be conservative in deciding an anomaly based on how small $\gamma$ is.

\section{Experiments}
In our experiments, we aim to answer two questions:
\begin{enumerate}
    \item Can our proposed method accurately identify anomalies in trajectory data?
    \item Is having an estimate of model uncertainty helpful in identifying anomalies?
\end{enumerate}
To answer (1), we evaluate our method on publicly available dataset and compare with baseline models based on standard classification metrics. To answer (2), instead of taking the normality score only, we incorporate the predictive variance when assigning anomalies which is discussed in Section \ref{sec:incorporateModelUncertainty}. We denote our proposed model with the uncertainty consideration as IRL-ADU (which stands for IRL Anomaly Detector with Uncertainty) and our model without the uncertainty consideration as IRL-AD. We compare these two methods with other baseline methods.

\subsection{Datasets}
We report on a comprehensive empirical study using real-world multidimensional  time series datasets which are publicly available: GeoLife GPS trajectory dataset \cite{zheng2010geolife,zheng2009mining} and Taxi Service Trajectory (TST) dataset \cite{moreira2013predicting}. The datasets do not include labeled anomalies.\footnote{To the best of our knowledge, there is no publicly available GPS trajectory datasets with labeled anomalies. Perhaps, the closest dataset is UAH-DriveSet \cite{romera2016need} which contains normal, aggressive, and drowsy driving behaviors with labels. However, the data size is very small with only 6 drivers and two trajectories per driver for each category.} Hence, we define what anomalies are in each dataset. 


 \textbf{GeoLife GPS Dataset} \cite{zheng2010geolife,zheng2009mining} contains GPS trajectories collected from 182 individual users over the period of three years and contains more than 17,000 trajectories. each trajectory contains the information of latitude, longitude and altitude. These trajectories were recorded by different GPS loggers and GPS-phones, and have a variety of sampling rates. 91 percent of the trajectories are logged every 1-5 seconds or every 5-10 meters per point. The length of trajectory varies significantly, ranging from below 100 to over 3000. The number of trajectories per user also varies from 28 to 400. This dataset recorded a broad range of users' outdoor movements, including not only life routines like go home and go to work but also some entertainments and sports activities, such as shopping, sightseeing, dining, hiking, and cycling. With this dataset, we chose 10 individuals (with larger number of trajectories recorded) to be our target agents, for each of whom we learn a separate reward function. We injected trajectories which do not belong to these individuals and define them as anomalies. Additionally, we hand-labeled some of the existing abnormal trajectories that are already present in those individuals trajectories (e.g. see the trajectory in the center in Figure \ref{fig:anomaly_scores})

\textbf{TST Dataset} \cite{moreira2013predicting} contains GPS trajectories collected from 442 taxi drivers in the city of Porto, in Portugal, and contains more than 1 million trips. Each trajectory contains the information of latitude and longitude, along with other meta information including how a service initiated (dispatch from central, taxi stand, or random streets), day type (workdays/weekends) and the location of an origin stand, etc. Number of trips per driver varies from 100 to 6751. Here, we are interested in learning the aggregated normal behavior of the taxi drivers; hence we learn a single reward function for the entire pool of drivers, and we define longer trips --- trips with cruise time greater than 50 minutes, which are mostly cross-town trips --- as anomalies. To prevent detection methods from being able to identify these anomalies by trivially looking at the length of the trips, we crop the anomalous trips from the end so that they are of similar lengths. Hence, the detection methods look at the earlier portion of these long trips and need to distinguish from other normal inner-city trips.

\subsubsection*{Data Preparation} In the evaluations on both datasets, we define state $s_t$ to be a vector of the location (longitude and latitude) at time $t$ concatenated with the initial location of the trajectory and the time duration since leaving the initial location. We define action $a_t$ to be the 2-dimensional vector representing a velocity at time $t$. One important thing to note is that a training dataset may also contain anomalous trajectories, however our model can deal with this sub-optimality since we adopt a Bayesian approach. In particular, this is crucial when deployed to real-world applications, since in practice we rarely have a clean scenario, free of any anomalies. Hence, for this reason, we can use our method in online settings where data arrive continuously and one can constantly update the reward function with a new stream of data. 

\subsection{Comparison with Other Methods}
As baselines, we consider two classic anomaly detection models: Local Outlier Factor (LOF) \cite{breunig2000lof} and One-class Support Vector Machines (One-Class SVM) \cite{manevitz2001one} to compare with our proposed model. Both baselines are state-of-the-art anomaly detection algorithms which can be used for sequential data \cite{gupta2014outlier}. The settings for both baselines are set to default of Scikit-learn \cite{pedregosa2011scikit} in Python library. Thus, both methods are able to return an anomaly score for each observation in a trajectory.
Additionally, we also compare with neural network autoencoder based detectors, similar to more recently work  \cite{malhotra2016lstm,oshea2016recurrent,sabokrou2018deep}. We consider both a feed-forward network autoencoder (FNN-AE) and a recurrent autoencoder, specifically Long short-term memory (LSTM-AE) model. The architecture and implementation details are deferred to the appendix.

\begin{figure*}[]
    \centering
    \includegraphics[width=0.93\linewidth]{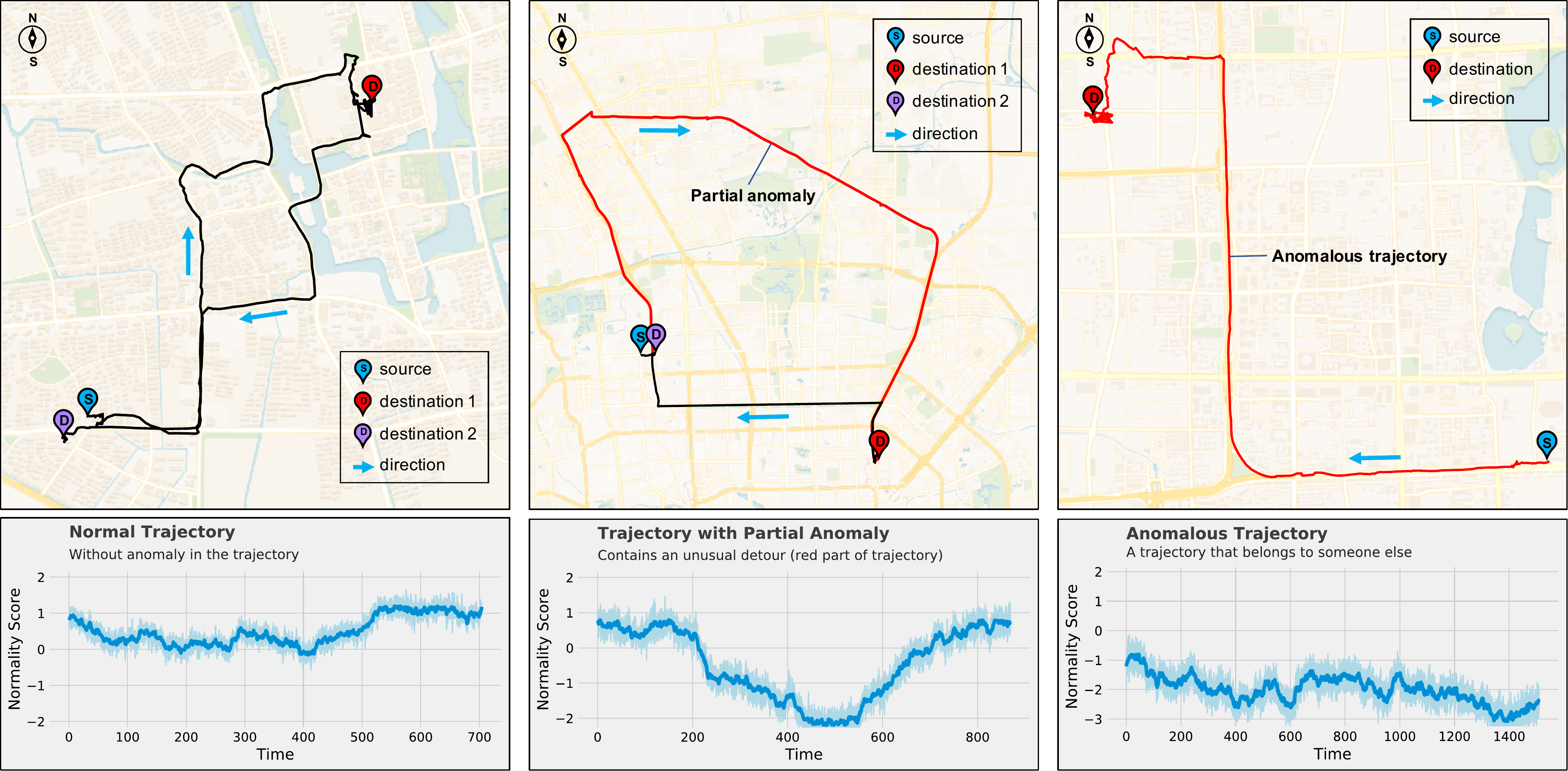}
    \caption{Sample trajectories from GeoLife GPS dataset and their corresponding predicted normality scores}
    \label{fig:anomaly_scores}
\end{figure*}


\subsection{Evaluation Metrics}
For evaluation, we compared predicted anomaly decisions to the ground truth, Boolean anomaly labels. Since the proposed method and baselines are able to identify anomaly observations, we are able to measure $F_1$-score, recall, and precision against the ground truth anomaly labels. The higher the values of $F_1$-score, recall, and precision are, the more accurate the given anomaly detection method is.

\subsection{Results}
We first discuss the sample results shown in Figure \ref{fig:anomaly_scores}. The left figure shows a normal trajectory which is a return trip from the source to Destination 1 and a return to Destination 2. The predicted normality scores suggest that the entire trajectory can be considered normal. Recall that the normality score is normalized in Eq. (\ref{eq:normalityScore}), hence we are only interested in negative deviations from zero when searching for anomalies --- and more positive normality scores represent potentially more frequently appearing trajectories than the average.
The center figure in Figure \ref{fig:anomaly_scores} shows a trajectory with a partial anomaly. This trajectory is one of the examples we chose to hand-label as anomaly after inspections due to the unusual detour 
which is highlighted as a red color trace on the figure. The predicted normality scores suggest the detour is highly likely an anomaly because of the low predicted normality score along that detour path.\footnote{Note that this does not mean all detours are anomalous. This trajectory happens to be a detour and anomalous with respect to normal behavior of the target agent.}
An interesting observation is that on the return path which is in fact a normal path, the normality score returns back to a normal range. 
This observation is very promising since it shows the proposed method can predict where/when the anomaly starts and where/wen it ends. This will be very useful when we deal with continuous streams of data sequences without the discrete notion of beginning and end of sequences. It suggests that our method can be used for real-time predictions. Lastly, the right figure shows the trajectory-wise anomaly which in fact belongs to someone else other than the target agent but was injected as the target agent's trajectory in the test time. Hence, this is an completely unobserved new sequence to the learned reward function. Hence, as expected, the predicted normality scores are very low for the entire duration. Note that on the third graph the y-axis scales are lower than the other two plots, and the scores are significantly lower than those of the other two trajectories; hence diagnosing this entire trajectory as an anomaly.

Based on the observations, we chose the threshold $\epsilon = -2$ and $\gamma = 1.5$. One could potentially fine-tune these threshold hyperparameters. However, for the evaluations we performed in this study, we fixed the threshold values. For the number of output heads in the reward network, we chose $K=10$ for all experiments.
Table \ref{tab:AvgPerformance} summarizes the average performance on the GeoLife GPS and TST datasets based on $F_1$, recall, and precision. Note that the performances on GeoLife GPS were averaged over the 10 users, for  whom we learn a separate reward function, and their detailed results are reported in Tables \ref{GeoLife-GPS-5} and \ref{GeoLife-GPS-10} in the appendix. For TST dataset, we learned a single reward function for the entire pool of the taxi drivers; hence learning a common behavior. 
The performance results show that both of our proposed methods, IRL-AD and IRL-ADU outperform other baseline methods. The results demonstrate that the proposed methods are effective at identifying outliers.
While these two methods show comparable results, IRL-ADU yields a higher precision overall and IRL-AD yield a slightly higher recall.

\vspace{-0.05cm}

\section{Discussion}
We propose a end-to-end framework for anomaly detection on sequential data. Our method utilizes IRL framework. 
To the best of our knowledge, this is the first attempt to use IRL framework in sequential anomaly detection problems. As stated earlier, this adaptation appears to be very natural and intuitive.
We believe this is an important result that shows viability of this line of work, which can lead further to bridging between the rich field of reinforcement learning and the practical application of anomaly detection. 

We also propose a scalable and simple technique to extend the state-of-art sample-based IRL method to approximate the posterior distribution of reward functions at a very low additional computational cost.  This extension may be of independent interest, which can also be used to solve a general IRL problems.

\vspace{-0.05cm}
\section{Acknowledgements}
The authors would like to thank Naoki Abe for helpful discussions.

\begin{table}
\begin{center}
\begin{tabular}{llcc}
    \toprule
Method &  & GeoLife & TST\\
    \midrule
\multirow{3}{*}{LOF} & Recall &	0.610&	0.483\\
                & Precision &	0.154&	0.478\\
                & $F_1$ score &	0.245&	0.481\\
    \midrule
\multirow{3}{*}{One-Class SVM}      & Recall &	0.620&	0.491\\
                                & Precision &	0.168&	0.474\\
                             & $F_1$ score &	0.263&	0.482\\
    \midrule
\multirow{3}{*}{FNN-AE} & Recall &	0.660&	0.521\\
                    & Precision &	0.173&	0.873\\
                 & $F_1$ score &	0.274&	0.652\\
    \midrule
\multirow{3}{*}{LSTM-AE} & Recall &	0.720&	0.555\\
                     & Precision &	0.199&	0.871\\
                    & $F_1$ score &	0.312&	0.678\\
    \midrule
\multirow{3}{*}{IRL-AD (Ours)} & Recall &	\textbf{0.850}&	\textbf{0.566}\\
                                    & Precision &	0.314&	0.924\\
                                    & $F_1$ score &	0.458&	\textbf{0.702}\\
    \midrule
\multirow{3}{*}{IRL-ADU (Ours)} & Recall &	0.840&	0.534\\
                    & Precision &	\textbf{0.353}&	\textbf{0.936}\\
                    & $F_1$ score &	\textbf{0.495}&	0.680\\
  \bottomrule
\end{tabular}
\end{center}
\caption{Average $F_1$, recall, and precision on real-world datasets and comparison with baselines}
\label{tab:AvgPerformance}
\end{table}

\vspace{-0.2cm}
%
\bibliographystyle{ACM-Reference-Format}
\bibliography{sample-base}

%

\clearpage
\appendix

\begin{table*}
\small
\begin{center}
\begin{tabular}{lcccccccccccccccccc}
    \toprule
&  \multicolumn{3}{c}{LOF} &  \multicolumn{3}{c}{One-class SVM} & \multicolumn{3}{c}{FNN-AE} &  \multicolumn{3}{c}{LSTM-AE} & \multicolumn{3}{c}{IRL-AD} & \multicolumn{3}{c}{IRL-ADU}\\
\cmidrule(lr){2-4}
\cmidrule(lr){5-7}
\cmidrule(lr){8-10}
\cmidrule(lr){11-13}
\cmidrule(lr){14-16}
\cmidrule(lr){17-19}
Data & R &  P & $F_1$ & R &  P & $F_1$ & R &  P & $F_1$ & R &  P & $F_1$ & R &  P & $F_1$ & R &  P & $F_1$\\
    \midrule
$U_1$ & 0.500&	0.071&	0.125&	0.500&	0.076&	0.132&	0.700&	0.100&	0.175&	0.700&	0.113&	0.194&	0.800&	0.148&	0.250&	0.800&	0.178&	0.291\\
$U_2$ & 0.700&	0.091&	0.161&	0.600&	0.078&	0.138&	0.800&	0.113&	0.198&	0.600&	0.120&	0.200&	0.700&	0.149&	0.246&	0.700&	0.167&	0.269\\
$U_3$ & 0.800&	0.095&	0.170&	0.800&	0.114&	0.200&	0.500&	0.082&	0.141&	0.700&	0.135&	0.226&	0.800&	0.131&	0.225&	0.800&	0.157&	0.262\\
$U_4$ & 0.600&	0.081&	0.143&	0.600&	0.072&	0.129&	0.600&	0.094&	0.162&	0.700&	0.113&	0.194&	0.800&	0.163&	0.271&	0.800&	0.157&	0.262\\
$U_5$ & 0.200&	0.037&	0.063&	0.300&	0.037&	0.065&	0.700&	0.104&	0.182&	0.800&	0.114&	0.200&	0.900&	0.170&	0.286&	0.900&	0.220&	0.352\\
$U_6$ & 0.600&	0.083&	0.146&	0.700&	0.119&	0.203&	0.500&	0.069&	0.122&	0.600&	0.086&	0.150&	0.800&	0.174&	0.286&	0.800&	0.182&	0.296\\
$U_7$ & 0.800&	0.098&	0.174&	0.700&	0.097&	0.171&	0.900&	0.127&	0.222&	0.900&	0.145&	0.250&	1.000&	0.175&	0.299&	1.000&	0.217&	0.357\\
$U_8$ & 0.600&	0.080&	0.141&	0.600&	0.074&	0.132&	0.700&	0.113&	0.194&	0.700&	0.101&	0.177&	0.800&	0.174&	0.286&	0.800&	0.174&	0.286\\
$U_9$ & 0.800&	0.118&	0.205&	0.500&	0.068&	0.120&	0.500&	0.094&	0.159&	0.600&	0.102&	0.174&	0.700&	0.140&	0.233&	0.700&	0.171&	0.275\\
$U_{10}$ & 0.600&	0.074&	0.132&	0.400&	0.058&	0.101&	0.600&	0.092&	0.160&	0.600&	0.090&	0.155&	0.800&	0.157&	0.262&	0.800&	0.160&	0.267\\
\midrule
Avg & 0.620&	0.083&	0.146&	0.570&	0.079&	0.139&	0.650&	0.099&	0.171&	0.690&	0.112&	0.192&	\textbf{0.810}&	0.158&	0.264&	\textbf{0.810}&	\textbf{0.178}&	\textbf{0.292}\\
 \bottomrule
\end{tabular}
\end{center}
\caption{Evaluation on GeoLife-GPS dataset with anomaly rate 5\%}
\label{GeoLife-GPS-5}
\end{table*}

\begin{table*}
\small
\begin{center}
\begin{tabular}{lcccccccccccccccccc}
    \toprule
&  \multicolumn{3}{c}{LOF} &  \multicolumn{3}{c}{One-class SVM} & \multicolumn{3}{c}{FNN-AE} &  \multicolumn{3}{c}{LSTM-AE} & \multicolumn{3}{c}{IRL-AD} & \multicolumn{3}{c}{IRL-ADU}\\
\cmidrule(lr){2-4}
\cmidrule(lr){5-7}
\cmidrule(lr){8-10}
\cmidrule(lr){11-13}
\cmidrule(lr){14-16}
\cmidrule(lr){17-19}
Data & R &  P & $F_1$ & R &  P & $F_1$ & R &  P & $F_1$ & R &  P & $F_1$ & R &  P & $F_1$ & R &  P & $F_1$\\
    \midrule
$U_1$ & 0.500&	0.119&	0.192&	0.500&	0.139&	0.217&	0.500&	0.125&	0.200&	0.600&	0.162&	0.255&	0.800&	0.296&	0.432&	0.800&	0.308&	0.444\\
$U_2$ & 0.500&	0.139&	0.217&	0.600&	0.146&	0.235&	0.600&	0.158&	0.250&	0.700&	0.179&	0.286&	0.900&	0.300&	0.450&	0.900&	0.346&	0.500\\
$U_3$ & 0.700&	0.194&	0.304&	0.700&	0.156&	0.255&	0.800&	0.186&	0.302&	0.800&	0.250&	0.381&	0.800&	0.364&	0.500&	0.800&	0.421&	0.551\\
$U_4$ & 0.700&	0.167&	0.269&	0.600&	0.162&	0.255&	0.700&	0.156&	0.255&	0.700&	0.171&	0.275&	0.900&	0.360&	0.514&	0.800&	0.348&	0.485\\
$U_5$ & 0.300&	0.086&	0.133&	0.500&	0.106&	0.175&	0.500&	0.139&	0.217&	0.500&	0.179&	0.263&	0.700&	0.259&	0.378&	0.700&	0.259&	0.378\\
$U_6$ & 0.600&	0.130&	0.214&	0.600&	0.182&	0.279&	0.500&	0.135&	0.213&	0.700&	0.194&	0.304&	0.800&	0.308&	0.444&	0.800&	0.381&	0.516\\
$U_7$ & 0.800&	0.211&	0.333&	0.700&	0.212&	0.326&	0.800&	0.250&	0.381&	0.900&	0.237&	0.375&	0.900&	0.310&	0.462&	0.900&	0.321&	0.474\\
$U_8$ & 0.700&	0.171&	0.275&	0.700&	0.212&	0.326&	0.800&	0.195&	0.314&	0.800&	0.222&	0.348&	0.900&	0.333&	0.486&	0.900&	0.429&	0.581\\
$U_9$ & 0.800&	0.195&	0.314&	0.800&	0.222&	0.348&	0.900&	0.243&	0.383&	0.900&	0.250&	0.391&	1.000&	0.370&	0.541&	1.000&	0.417&	0.588\\
$U_{10}$ & 0.500&	0.125&	0.200&	0.500&	0.139&	0.217&	0.500&	0.143&	0.222&	0.600&	0.150&	0.240&	0.800&	0.242&	0.372&	0.800&	0.296&	0.432\\
\midrule
Avg & 0.610&	0.154&	0.245&	0.620&	0.168&	0.263&	0.660&	0.173&	0.274&	0.720&	0.199&	0.312&	\textbf{0.850}&	0.314&	0.458&	0.840&	\textbf{0.353}&	\textbf{0.495}\\
 \bottomrule
\end{tabular}
\end{center}
\caption{Evaluation on GeoLife-GPS dataset with anomaly rate 10\%}
\label{GeoLife-GPS-10}
\end{table*}



\section{Neural Network Details}
FNN-AE is a fully-connected feed forward autoencoder with 3 hidden layers  with dimensions 64, 16, and 64. We use Relu activation functions for all hidden layers and linear activation at the output. LSTM-AE  is a hierarchical model that consists of an encoder and a decoder, each of which is an LSTM. For our reward function, we use two hidden layers of 64 and 16, the network branches out independently to $K$ fully connected output nodes. In all experiments, we used $K = 10$. For policy optimization, we use TRPO method \cite{schulman2015trust} for which we used openAI baseline model.

\section{Per User Experiments on GeoLife-GPS}
We picked 10 users with largest number of trajectories. Among them, we filtered trajectories that are shorter than 100 time-steps. After training, for each user, we injected (and/or identified) anomalies so that the total proportions of anomalies in the test datasets are 5\% (Table \ref{GeoLife-GPS-5}) and 10\% (Table \ref{GeoLife-GPS-10}) respectively in each independent experiments. The results are shown in Tables \ref{GeoLife-GPS-5} and \ref{GeoLife-GPS-10}.

\section{Bayesian neural network}\label{sec:BNN}

In this section, we briefly discuss the work on Bayesian neural network.
Let $\mathcal{D} = \{ (\bm{x}_i, \bm{y}_i)\}^N_{i=1}$ be a collection of realizations of i.i.d random variables, where $\bm{x}_i$ is an image, $\bm{y}_i$ is a corresponding density map, and $N$ denotes the sample size. In Bayesian neural network framework, rather than thinking of the weights of the network as fixed parameters to be optimized over, it treats them as random variables, and so we place a prior distribution $p(\theta)$ over the weights of the network $\theta \in \theta$. This results in the posterior distribution
\begin{equation*}
    p(\theta | \mathcal{D}) = \frac{p(\mathcal{D} | \theta)p(\theta)}{p(\mathcal{D})} = \frac{\left(\prod^N_{i=1} p(y_i | x_i, \theta)\right) p(\theta)}{p(\mathcal{D})}.
\end{equation*}
While this formalization is simple, the learning is often challenging because calculating the posterior $p(\theta | \mathcal{D})$ requires an integration with respect to the entire parameter space $\Theta$ for which a closed form often does not exist. \cite{mackay1992practical} proposed a Laplace approximation of the posterior.
\cite{neal1993bayesian} introduced the Hamiltonian Monte Carlo, a Markov Chain Monte Carlo (MCMC) sampling approach using Hamiltonian dynamics, to learn Bayesian neural networks. This yields a principled set of posterior samples without direct calculation of the posterior but it is computationally prohibitive. 
Another Bayesian method is variational inference \cite{blundell2015weight, graves2011practical,  louizos2017multiplicative} which approximates the posterior distribution by a tractable variational distribution $q_\eta(\theta)$ indexed by a variational parameter $\eta$. The optimal variational distribution is the closest distribution to the posterior among the pre-determined family $Q = \{q_\eta(\theta)\}$. The closeness is often measured by the Kullback-Leibler (KL) divergence between $q_\eta(\theta)$ and $p(\theta | \mathcal{D})$. While these Bayesian neural networks are the state of art at estimating predictive uncertainty, these require significant modifications to the training procedure and are computationally expensive compared to standard (non-Bayesian) neural networks 

\cite{gal2016dropout} proposed using Monte Carlo dropout to estimate predictive uncertainty by using dropout at test time. There has been work on approximate Bayesian interpretation of dropout \cite{gal2016dropout, kingma2015variational}. Specifically, \cite{gal2016dropout} showed that Monte Carlo dropout is equivalent to a variational approximation in a Bayesian neural network. With this justification, they proposed a method to estimate predictive uncertainty through variational distribution. Monte Carlo dropout is relatively simple to implement leading to its popularity in practice. Interestingly, dropout may also be interpreted as ensemble model combination \cite{srivastava2014dropout} where the predictions are averaged over an ensemble of neural networks. The ensemble interpretation seems more plausible particularly in the scenario where the dropout rates are not tuned based on the training data, since any sensible approximation to the true Bayesian posterior distribution has to depend on the training data. This interpretation motivates the investigation of ensembles as an alternative solution for estimating predictive uncertainty. Despite the simplicity of dropout implementation, we were not able to produce satisfying confidence interval for our crowd counting problem. Hence we consider a simple non-parametric bootstrap of functions which we discuss in the following section.

\end{document}